\documentclass[letterpaper, 10 pt, conference]{ieeeconf}  

\IEEEoverridecommandlockouts                              

\usepackage{xcolor}
\usepackage{colortbl} 
\usepackage{multirow}
\usepackage{graphics} 
\usepackage{epsfig} 
\usepackage{mathptmx} 
\usepackage{times} 
\usepackage{amsmath} 
\usepackage{amssymb}  
\usepackage{balance}
\usepackage[colorlinks=true,
            linkcolor=blue,
            urlcolor=blue,
            citecolor=blue]{hyperref}

\title{\LARGE \bf
All-St3R SLAM: Real-time RGB Dynamic SLAM with Pixel-wise Motion Masking
}

\author{
    Yewon Kim$^{1}$, Paul Hongsuck Seo$^{2}$, Sang Chul Ahn$^{3\dagger}$%
    \thanks{$^\dagger$ \parbox[t]{0.9\columnwidth}{Corresponding author}}%
    \thanks{$^{1}$ \parbox[t]{0.9\columnwidth}{Yewon Kim is with Center for Artificial Intelligence Research, Korea Institute of Science and Technology, Seoul, Korea (South)
            {\tt\small yewon@kist.re.kr}}}%
    \thanks{$^{2}$ \parbox[t]{0.9\columnwidth}{Paul Hongsuck Seo is with Dept. of Computer Science and Engineering, Korea University, Seoul, Korea (South)
            {\tt\small phseo@korea.ac.kr}}}%
    \thanks{$^{3}$ \parbox[t]{0.9\columnwidth}{Sang Chul Ahn is with Center for Artificial Intelligence Research, Korea Institute of Science and Technology, and with AI-Robotics, KIST School (UST), Seoul, Korea (South)
            {\tt\small asc@kist.re.kr}}}%
}

\begin{document}

\maketitle
\thispagestyle{empty}
\pagestyle{empty}


\begin{abstract}

SLAM (Simultaneous Localization and Mapping) is a core technology in autonomous systems that enables them to estimate their own position and build a map of unknown environments simultaneously. 
SLAM systems in dynamic environments are of great practical importance. However, existing dynamic SLAM approaches still have several limitations
like over-removal and high computational costs. 
To address these limitations, we propose a real-time RGB Dynamic SLAM framework that precisely removes moving objects at the pixel level without relying on predefined labels or segmentation models. Our system combines pixel motion predicted by ego-motion using DUSt3R with point tracking from All-Tracker, removing only actually moving pixels by analyzing discrepancies between the two motion sources.
Unlike object-level masking methods, our approach eliminates only currently moving regions, allowing temporarily static parts to be utilized for tracking. 
Operating at approximately 10 FPS, the system achieves real-time performance on standard 30 FPS video streams by subsampling every 3 frames. Relying solely on RGB inputs, this significantly improves upon state-of-the-art dense dynamic SLAM approaches (0.5–3 FPS), presenting a highly practical solution. Project page: \url{https://yellowon0103.github.io/all-st3r-slam/}

\end{abstract}


\section{INTRODUCTION}
For embodied AI systems, such as service robots, autonomous drones, and mobile manipulators, Simultaneous Localization and Mapping (SLAM) has become a fundamental perception module in real-world applications \cite{cadena2017past,duan2022survey}. 
However, real-world environments often contain moving objects, motivating the development of SLAM systems that can operate robustly even in dynamic environments.


\definecolor{myyellow}{RGB}{255,220,0}
\definecolor{mygreen}{RGB}{0,231,0}
\begin{figure}[t]
    \centering
    \includegraphics[width=0.7\columnwidth]{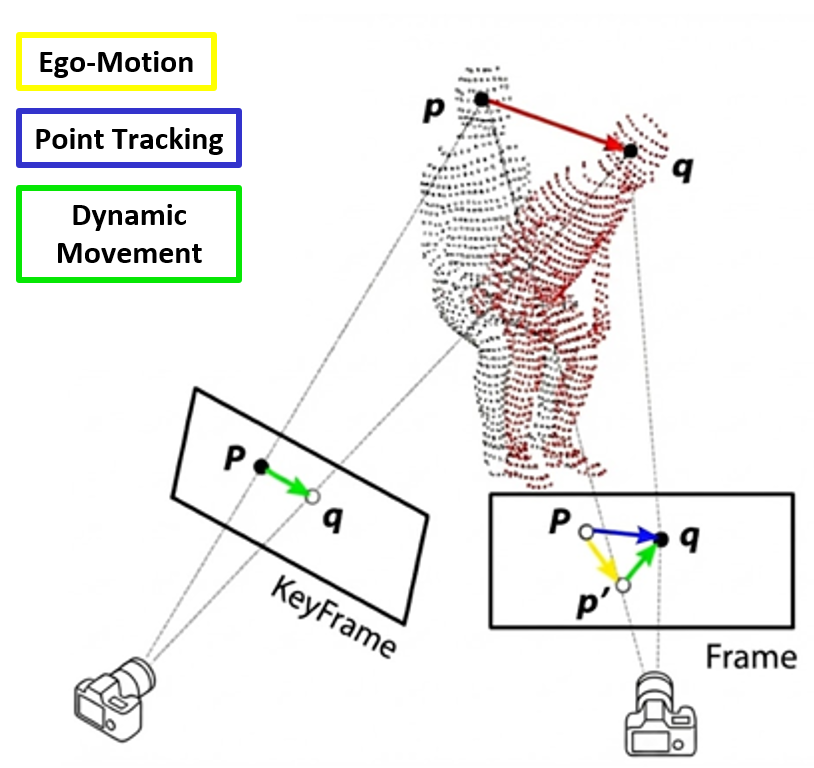}
    \caption{
    Comparison between \textcolor{myyellow}{Ego-Motion}, 
    \textcolor{blue}{Point Tracking}, and 
    \textcolor{mygreen}{Dynamic Movement}. 
    Assuming the KeyFrame point cloud reconstructed by DUSt3R is static, 
    projecting those 3D points into the Frame camera produces 
    the pixel motion predicted by camera
    \textcolor{myyellow}{Ego-Motion}. 
    In contrast, All-Tracker predicts pixel correspondences directly 
    between the KeyFrame and the Frame in the image domain, producing 
    \textcolor{blue}{Point Tracking}. 
    When the KeyFrame point cloud contains dynamic objects, 
    discrepancies arise between these two motions. 
    Points with large discrepancies between the two motions 
    are regarded as \textcolor{mygreen}{Dynamic Movement}.
    }
    \label{fig:sceneflow}
    \vspace{-20pt}
\end{figure}

Recent dynamic SLAM research has proposed two main strategies:
(1) joint tracking of camera and dynamic objects, and (2) dynamic removal. Although tracking-based methods \cite{zhang2020vdo, yang2019cubeslam} provide comprehensive scene-level understanding, they are often too computationally expensive. 
Consequently, the removal strategy—masking dynamic regions—has become the standard practical approach.

Existing removal-based approaches, however, face significant trade-offs. Early Geometric methods \cite{tan2013robust} are lightweight but remain sparse, while object-level and hybrid approaches \cite{bescos2018dynaslam,long2025dog,zheng2025wildgs} rely on heavy segmentation or detection priors. These methods not only suffer from over-removal (e.g., discarding parked cars) and limited generalization to unseen categories, but also introduce significant computational overhead, restricting real-time deployment.

Thus, we propose a dense, geometry-driven dynamic RGB SLAM system capable of per-pixel motion masking that achieves real-time performance without relying on heavy object-level segmentation or detection.
We integrate dense 3D matching from DUSt3R \cite{wang2024dust3r} with lightweight pairwise point tracking adapted from All-Tracker \cite{harley2025alltracker}.
As illustrated in Fig.~\ref{fig:sceneflow}, dynamic regions are identified by comparing the pixel motion predicted by ego-motion with the observed point tracking motion. 
Our system offers several crucial advantages:

\begin{itemize}

\item \textbf{Dynamic removal without segmentation or detection models:} 
We detect dynamic regions purely from geometric discrepancy without relying on object-level segmentation or detection networks. This label-free formulation generalizes to unseen objects and preserves static pixels within partially moving objects, significantly reducing over-removal.

\item \textbf{Robust tracking via progressive dynamic masking:} 
Unlike single-step removal strategies that risk discarding useful correspondences, our method progressively accumulates motion inconsistencies across frames. This strategy enables preserving maximal tracking correspondences, leading to robust tracking in dynamic scenes.

\item \textbf{Real-time RGB-only framework:} 
By adapting All-Tracker to a pairwise two-frame inference setting, we eliminate sequential overhead without sacrificing tracking quality. Combined with lightweight geometric filtering, our system operates at approximately 10 FPS. By subsampling every 3 frames, this throughput enables real-time performance on standard 30 FPS video streams.

\end{itemize}

Our approach prioritizes real-time performance and high tracking accuracy using solely monocular RGB inputs. 


\section{RELATED WORKS}


Recent research in dynamic SLAM has developed two main strategies: estimating the motion of objects (Tracking) and identifying dynamic objects as outliers (Removal).
While joint tracking (e.g., VDO-SLAM \cite{zhang2020vdo} and CubeSLAM \cite{yang2019cubeslam}) enables comprehensive scene-level understanding, it incurs substantial computational costs and struggles with non-rigid or unstructured motion. We focus on the removal-based approaches, which 
can be broadly grouped into
three paradigms: geometric, object-level masking, and hybrid.

Early \textbf{Geometric} methods rely purely on motion cues—identifying inconsistencies across views using epipolar geometry, point tracking residuals, or depth disagreement \cite{tan2013robust}. For example, Tan et al. \cite{tan2013robust} detect dynamic points by evaluating their inconsistency with the dominant scene motion. These early feature-based techniques are lightweight and training-free, but 
mostly sparse.

\textbf{Object-level masking} methods employ segmentation networks (e.g., Mask R-CNN), object detectors (e.g., YOLO), or semantic-rich embeddings (e.g., DINOv2 \cite{oquab2023dinov2}) to mask potentially dynamic object classes. DynaSLAM \cite{bescos2018dynaslam} removes all features belonging to categories that are potentially dynamic, such as people or vehicles. However, this approach leads to systematic over-removal: 
stationary people or parked cars are discarded.
Additionally, semantic segmentation or detection inference carries high computational costs. DynaSLAM 
operates significantly below real-time performance (typically 1–2 FPS) \cite{bescos2018dynaslam}. Furthermore, these approaches 
may have difficulty generalizing
to novel object categories outside their training vocabulary (e.g., COCO-80 \cite{lin2014microsoft}), such as electric scooters or delivery robots.

\textbf{Hybrid} methods integrate object-level masking with geometric motion verification to overcome the core limitations of purely object-level masking methods. Rather than removing all instances of potentially dynamic classes, hybrid systems like DOG-SLAM \cite{long2025dog} first detect candidate objects and then verify geometric consistency to identify genuinely dynamic elements. WildGS-SLAM \cite{zheng2025wildgs} 
uses semantic-rich feature embeddings as an implicit prior for uncertainty estimation
and 3D Gaussian Splatting \cite{kerbl20233d} for photometric-geometric consistency checking. It demonstrates strong trajectory accuracy on benchmark datasets but suffers from severe computational bottlenecks. 

\begin{figure*}[t]  
\begin{center}
\includegraphics[width=\textwidth]{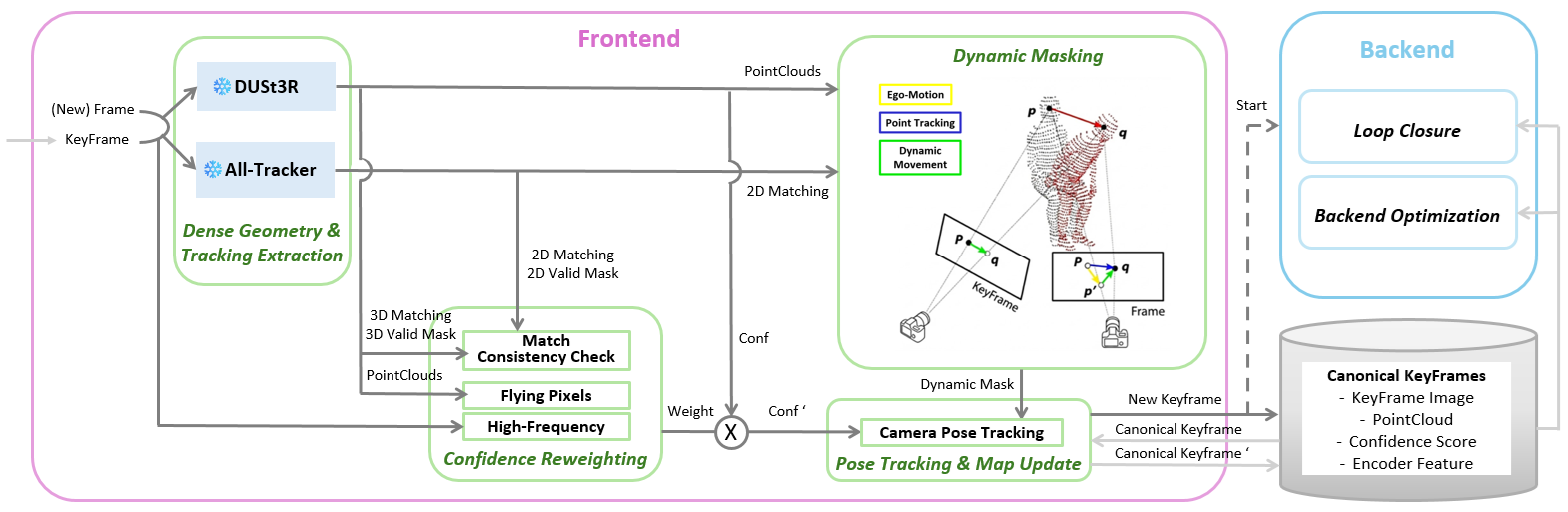} 
\end{center}
\caption{Overview of the proposed Monocular RGB Dynamic SLAM system.}
\label{fig:architecture}
\vspace{-15pt}
\end{figure*}






\section{Methodology}

Our proposed framework is a real-time monocular dense SLAM system 
designed for robust tracking and mapping in dynamic environments. 
We build upon MASt3R-SLAM~\cite{murai2025mast3r} as our baseline, but introduce a fundamental shift in how geometric correspondences are constructed and dynamic 
outliers are handled.

We adopt DUSt3R~\cite{wang2024dust3r} as our core geometric backbone 
in place of MASt3R, benefiting from its faster inference and lower 
memory footprint. To compensate for the absence of MASt3R's learned 
matching features, we integrate a lightweight pairwise adaptation of 
All-Tracker~\cite{harley2025alltracker}, which supplies dense 
image-level point tracking as a complementary correspondence source. 
Crucially, this combination serves a dual purpose: beyond providing high-fidelity correspondences, it enables purely geometry-driven dynamic masking without the need for any segmentation or detection models.

\subsection{System Overview}

To clarify our pipeline, we define 
three core data terms:

\begin{itemize}
    \item \textbf{Frame ($I_f$):} The current RGB input. 

    \item \textbf{Keyframe ($I_k$):} A historically significant 
    frame used as a reference for tracking and backend graph 
    optimization. 

    \item \textbf{Canonical Keyframe Pointcloud 
    ($\tilde{X}_k^k$):} A 3D pointmap representation of a keyframe defined in the 
    global coordinate frame, initially generated via DUSt3R 
    self-coupling inference ($I_k, I_k$). Each keyframe maintains 
    its own independent pointcloud.
\end{itemize}

The pipeline processes each incoming frame $I_f$ by pairing it with 
the current keyframe $I_k$. As illustrated in 
Fig.~\ref{fig:architecture}, the system proceeds through four 
sequential stages.

\textbf{Dense Geometry and Tracking Extraction.}
$I_f$ and $I_k$ are jointly fed into DUSt3R, which outputs two 
aligned dense pointclouds—the frame pointcloud $X_f^f$ and the 
keyframe pointcloud $X_f^k$—along with per-point geometric 
confidence scores and 3D matching. In parallel, All-Tracker 
processes the same image pair to produce dense point tracking(2d matching) with associated confidence and visibility scores.

\textbf{Confidence Reweighting.}
Raw confidence scores are refined based on 2D-3D consistency, high-frequency textures, and depth boundary filtering to suppress noisy geometries and enhance discriminative anchors for subsequent tracking stages. (Section~\ref{sec:Reweighting})

\textbf{Pixel-wise Dynamic Masking.}
Independent motions are filtered at the pixel level by computing the discrepancy between ego-motion predictions and All-Tracker point tracking. This progressive frame-by-frame masking isolates moving outliers while preserving valid static points within dynamic entities. (Section~\ref{sec:Dynamic_Masking})

\textbf{Pose Tracking and Map Update.}
The refined, dynamic-free correspondence set is used to estimate 
the camera $\text{Sim}(3)$ pose of $I_f$ with respect to the 
canonical keyframe pointcloud $\tilde{X}_k^k$ via weighted 
optimization. If the number of valid correspondences falls below 
a threshold, the current frame is promoted to a new keyframe. When a new keyframe is created, 
the backend performs loop closure detection and global pose graph 
optimization to maintain a globally consistent 3D map.


\subsection{Confidence Reweighting}
\label{sec:Reweighting}

Confidence Reweighting adjusts the importance of pixel correspondences before tracking, as accurate camera pose estimation relies heavily on precise pixel-wise correspondences between the current Frame and the reference Keyframe. Following the optimization formulation of MASt3R-SLAM~\cite{murai2025mast3r}, these confidence scores serve directly as weights to guide the optimizer toward high-confidence matches. Unlike MASt3R-SLAM, we propose an enhanced correspondence generation strategy that fuses geometric matching from DUSt3R~\cite{wang2024dust3r} with robust point tracking from All-Tracker~\cite{harley2025alltracker}. By combining geometric consistency with visual tracking cues, this fusion provides confidence-aware weighting that allows the optimizer to prioritize the most trustworthy matches for pose estimation.



First, we utilize DUSt3R~\cite{wang2024dust3r} to predict dense pointmaps for the input image pair ($I_k, I_f$). Specifically, we obtain $X_f^f$, the pointmap of the current frame in its own coordinate system, and $X_f^k$, the pointmap of the keyframe image expressed in the current frame's camera coordinate system.

To establish correspondences from these pointmaps, we adopt the iterative projective matching method proposed in MASt3R-SLAM~\cite{murai2025mast3r}. 
Iterative projective matching yields a set of dense geometric matches, denoted as $\mathcal{M}_{geo}$, which are derived purely from the 3D geometric priors provided by DUSt3R~\cite{wang2024dust3r}. In addition, we obtain point tracking correspondences between $I_k$ and $I_f$ using All-Tracker, which are denoted as $\mathcal{M}_{flow}$. 

We then perform a consistency check between geometric matches $\mathcal{M}_{geo}$ and point tracking matches $\mathcal{M}_{flow}$. For a given pixel $u$ in the keyframe, if the correspondence predicted by geometric matching lies within a small radius of the correspondence predicted by point tracking, the match is verified as
\vspace{-4pt}
\begin{equation}
    \| \mathcal{M}_{geo}(u) - \mathcal{M}_{flow}(u) \|_2 < \delta_{pixel}.
\end{equation}
\vspace{-4pt}
$\delta_{pixel}$ denotes a pixel consistency threshold that defines the maximum allowable distance between the two correspondences (set to 2 pixels in our implementation).
Matches that satisfy this condition are considered to be both geometrically valid and visually consistent. For these verified matches, we increase their associated confidence scores (originally from DUSt3R~\cite{wang2024dust3r}) by a scalar factor $\gamma > 1.0$ (empirically set to 1.2 in our implementation). This weighting strategy ensures that these highly reliable correspondences exert a stronger influence during the subsequent pose optimization phase.

Beyond match-level verification, we further refine correspondence reliability through pixel-level geometric and photometric analysis. Dense depth estimation models, including DUSt3R~\cite{wang2024dust3r}, often produce artifacts known as flying pixels near object boundaries. These artifacts arise from depth discontinuities and introduce erroneous 3D geometry, which degrades correspondence reliability. To mitigate this issue, we compute the normalized Laplacian of the depth map to detect abrupt depth variations. Pixels with large normalized responses are identified as flying pixels, and their influence in pose optimization is suppressed through reduced confidence weighting. This geometry-aware filtering limits the impact of noisy boundary regions during camera pose estimation.

Conversely, to prioritize reliable tracking regions, we apply a Laplacian operator directly to the RGB images ($I_k$ and $I_f$) to extract high-frequency textural details. Pixels exhibiting strong edge responses represent distinct, traceable visual features. We apply a confidence boost to these high-frequency regions. 


\subsection{Dynamic Masking}
\label{sec:Dynamic_Masking}
Our system identifies dynamic regions at the pixel level by analyzing the discrepancy between the pixel motion predicted by ego-motion and the observed point tracking motion. 

We formulate the problem under a static world assumption where pixel motion between the Keyframe and the current Frame is governed solely by the camera's relative pose and the scene geometry.
First, we estimate the relative pose $T_{rel} \in \text{SE}(3)$ between the Keyframe pointcloud $X_f^k$ and the current Frame pointcloud $X_f^f$. Since DUSt3R~\cite{wang2024dust3r} outputs dense pointmaps for both images that are intrinsically aligned in the Frame's coordinate system, and given the one-to-one correspondence between 2D pixels and 3D points, we can straightforwardly solve for the camera extrinsic parameters using the PnP-RANSAC method. 
This formulation is robust against the circular dependency between ego-motion estimation and dynamic detection. Since DUSt3R implicitly learns rigid view transformations through its cross-attention layers, it naturally assigns lower confidence scores to dynamic objects that violate epipolar geometry constraints~\cite{chen2025easi3r}. Consequently, the network outputs point clouds by focusing the alignment on the static background, ensuring a robust relative pose estimation even in dynamic environments, as shown in Section~\ref{sec:Experiments}.

Assuming all points in the Keyframe are static, we project each 3D point $P_k$ from the Keyframe $X_f^k$ onto the image plane of the current Frame using the estimated relative pose $T_{rel}$ and intrinsic matrix $K$.
The predicted ego-motion flow vector $F_{ego}(u_k)$ for a pixel $u_k$ is defined as the displacement induced by the camera motion,
\vspace{-2pt}
\begin{equation}
    F_{ego}(u_k) = u_{proj} - u_k,
\vspace{-2pt}
\end{equation}
which represents the expected motion of static structures in the scene under camera ego-motion.
Simultaneously, we obtain the Point Tracking from All-Tracker~\cite{harley2025alltracker}, which captures the actual visual displacement of pixels based on image appearance. Let $u_{track}$ be the pixel location in $I_f$ corresponding to $u_k$ as tracked by All-Tracker. The point tracking vector $F_{pointTracking}(u_k)$ is defined as:
\begin{equation}
    F_{pointTracking}(u_k) = u_{track} - u_k,
\end{equation}
which reflects the true motion of pixels, including the movement of dynamic objects.
For static points, the ego-motion predicted flow and point tracking motion should coincide (i.e., $F_{ego} \approx F_{pointTracking}$). However, for dynamic objects, a discrepancy arises because the object's independent motion is not accounted for in the ego-motion prediction, as illustrated in Fig.~\ref{fig:sceneflow}. We define the flow discrepancy $\Delta F(u_k)$ as the Euclidean distance between the two flow vectors,
\begin{equation}
    \Delta F(u_k) = \| F_{ego}(u_k) - F_{pointTracking}(u_k) \|_2,
\end{equation}
where pixels with large discrepancies indicate independent motion.
Pixels where $\Delta F(u_k)$ exceeds a dynamic threshold $\tau_{dyn}$ are classified as dynamic, where $\tau_{dyn}$ is set to seven times the median absolute deviation in $\Delta F(u_k)$.
This check effectively isolates independent motion (e.g., a moving arm) while preserving the static parts of an object (e.g., the human torso), maximizing the number of usable tracking features.

\begin{figure}
\begin{center}
\includegraphics[width=\columnwidth]{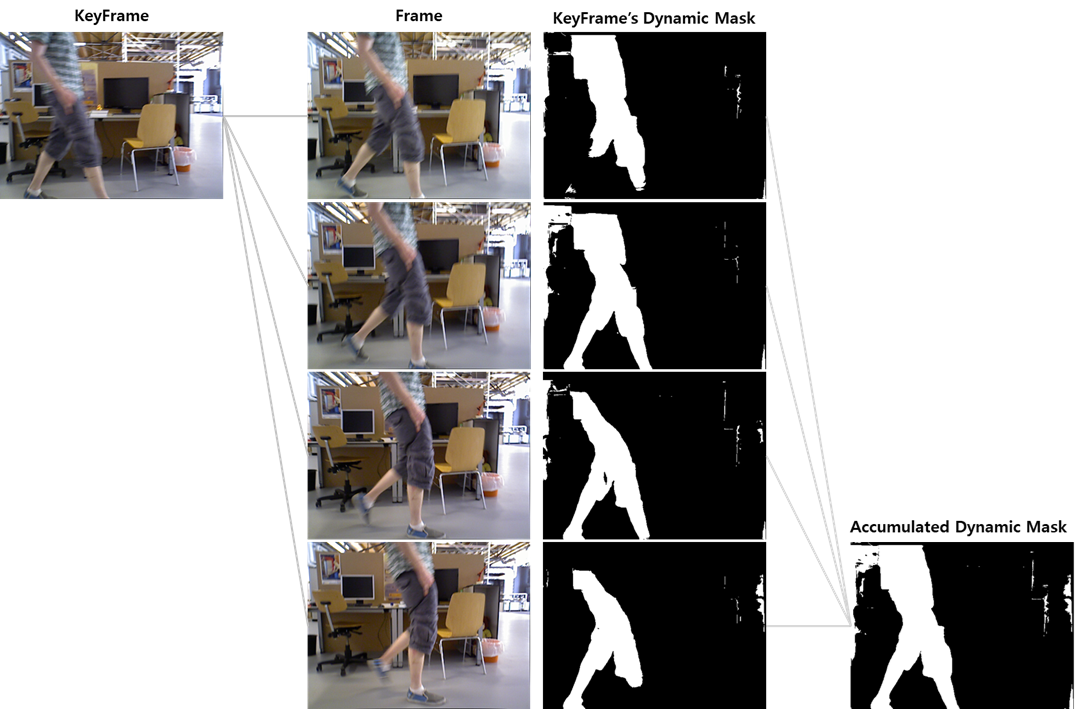}
\end{center}
\caption{Illustration of the frame-by-frame mask accumulation.}
\label{fig:mask_accumulation}
\vspace{-15pt}
\end{figure}

To handle dynamic objects robustly over time, we employ a frame-by-frame mask accumulation strategy (Fig.~\ref{fig:mask_accumulation}). For the current Keyframe $I_k$, we maintain a cumulative dynamic mask $Dyn_{k}$. When a new Frame $I_f$ arrives at time step $t$, the dynamic mask $dyn_{t}$ is computed based on the motion discrepancy $\Delta F$ between $I_k$ and $I_f$, and then aggregated via a union operation: $Dyn_{k} \leftarrow Dyn_{k} \cup dyn_{t}$. This process repeats for all subsequent frames arriving until a new Keyframe is selected, accumulating all dynamic regions observed relative to $I_k$. This accumulation ensures that any dynamic region identified at any point remains masked, preventing transient static observations of moving objects from corrupting the map. 

Furthermore, we employ a moderate frame subsampling strategy that not only enables real-time performance but also improves dynamic region removal. By increasing the temporal baseline between $I_k$ and $I_f$, it magnifies the pixel displacements of slow-moving objects. This increased displacement makes subtle motion more distinguishable from the static background, allowing the dynamic masking module to more reliably identify and remove slowly moving regions that might otherwise be regarded as static.

\subsection{Lightweight Two-Frame All-Tracker}
The original All-Tracker is designed for offline processing of long video clips, relying on computationally intensive temporal attention mechanisms across large sliding windows. This design induces high latency and memory usage, making it unsuitable for real-time SLAM.
To address this, we reconfigured the network architecture for lightweight pairwise inference. We bypassed the long-range temporal attention modules, constraining the model to accept only adjacent frame pairs $(I_{i}, I_{i+1})$ as input. This modification significantly reduces the computational graph and inference time while retaining the model's superior capability in handling fine-grained pixel displacements.


\section{Experiments}
\label{sec:Experiments}

We evaluated the performance of the proposed SLAM system through extensive experiments on multiple dynamic and static datasets. 
All experiments were conducted on a single NVIDIA RTX 4090 GPU. Our system was implemented in PyTorch. For all input images, we resized them such that the maximum dimension was 512 pixels. To achieve effective real-time performance, we processed the input sequences with a 3-frame subsampling strategy. 

\begin{figure}[t]
\centering
\includegraphics[width=\columnwidth]{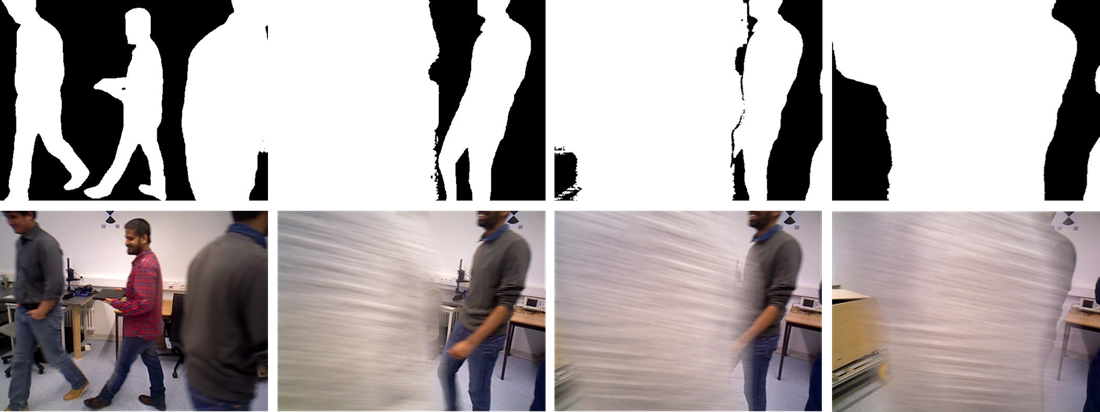}
\caption{\textbf{Object-Level Masking on Bonn Dataset (SAM2).} Visualization of dynamic masks generated by SAM2 \cite{ravi2024sam2} on the crowd sequence of the Bonn RGB-D Dataset. Even when motion is limited to a specific part (e.g., moving arms), the entire human body is masked out. This over-removal discards valid static regions such as the torso, significantly reducing the number of valid matching correspondences required for stable tracking.}
\label{fig:sam2_mask}
\vspace{-3pt}
\end{figure}

\begin{figure}[t]
\centering
\includegraphics[width=\columnwidth]{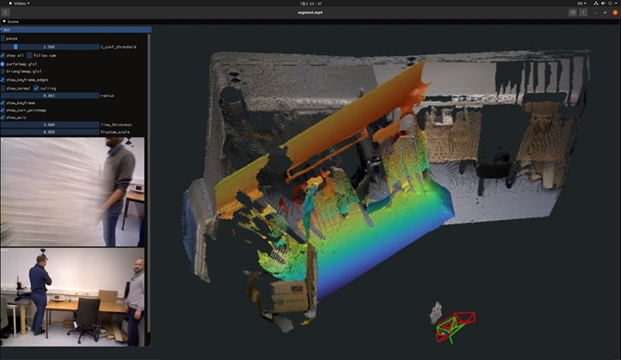}
\caption{\textbf{Tracking Failure with Object-Level Masking.}
The visualization shows the system status using SAM2 masks.\\}
\label{fig:segmentation_not_stable}

\includegraphics[width=\columnwidth]{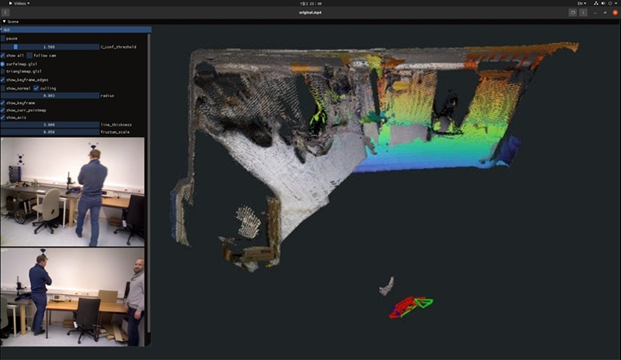}
\caption{\textbf{Stable Tracking with Pixel-Wise Motion Masking (Ours).}
Under the same dynamic conditions, our method selectively masks only moving parts while preserving static regions.
The system maintains continuous tracking without any relocalization. This confirms that utilizing temporarily static features on dynamic objects is helpful for robustness.}
\label{fig:ours_stable}
\vspace{-15pt}
\end{figure}

\subsection{Robust tracking stability}
\begin{figure}
    \centering
    \begin{minipage}{0.49\columnwidth}
        \centering
        \includegraphics[width=\columnwidth]{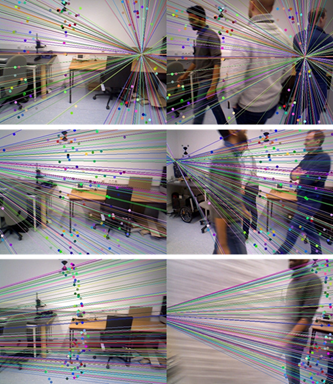}
    \end{minipage}
    \hfill
    \begin{minipage}{0.49\columnwidth}
        \centering
        \includegraphics[width=\columnwidth]{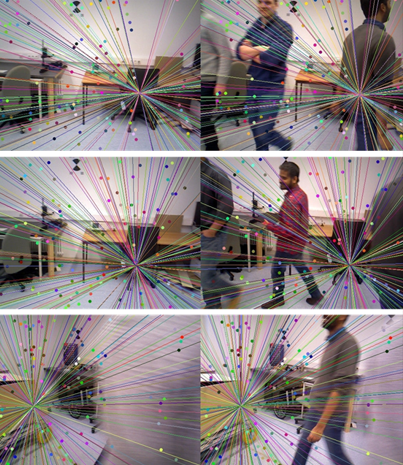}
    \end{minipage}
    
    \caption{\textbf{Geometric Verification under Dynamic Occlusion.}
    \textbf{(Left: Keyframe, Right: Current Frame).}
    Visualization of valid matches (dots) and epipolar lines. Despite large dynamic obstructions, the system maintains accurate epipolar constraints and secures a sufficient number of reliable matches, confirming stable pose estimation.}
    \label{fig:epipolar_verification}
    \vspace{-15pt}
\end{figure}

We first evaluate our core claim that frame-by-frame pixel-wise motion masking improves tracking stability in highly dynamic scenes by preserving valid correspondences discarded by object-level masking. Unlike long-term trajectory accuracy, tracking stability reflects continuous online pose estimation.
As shown in Fig.~\ref{fig:sam2_mask}, object-level masking via SAM2~\cite{ravi2024sam2} causes over-removal by discarding temporarily static regions 
leading to severe feature starvation, and consequently triggering three relocalizations within just 20 keyframes (Fig.~\ref{fig:segmentation_not_stable}). In contrast, our pixel-wise masking selectively removes only moving pixels, preserving static components as valid geometric anchors. This maintains stable tracking without relocalization throughout the sequence (Fig.~\ref{fig:ours_stable}). The rainbow point cloud of the current frame aligns perfectly with the global map, confirming the high precision of the estimated pose. Furthermore, Fig.~\ref{fig:epipolar_verification} shows that our method preserves accurate epipolar constraints even under severe dynamic occlusions.

\subsection{Speed with High Tracking Accuracy}
Table~\ref{tab:runtime_accuracy} presents the comparison of Runtime (FPS) and Absolute Trajectory Error (ATE RMSE) on the Wild-SLAM and Bonn datasets.
Our method achieves \textbf{$\sim$11.24 FPS} on Wild-SLAM and \textbf{$\sim$10.13 FPS} on Bonn with peak GPU memory usages of 7.65 GB and 9.46 GB, respectively. This is significantly faster than the SOTA method, WildGS-SLAM \cite{zheng2025wildgs}, which operates at roughly 0.5 FPS (Full) to 2 FPS (Fast).
Despite this high speed, our tracking accuracy is comparable to WildGS-SLAM \cite{zheng2025wildgs}.

\begin{table}
\centering
\caption{Runtime (FPS $\uparrow$) and ATE RMSE (cm $\downarrow$).}
\label{tab:runtime_accuracy}
\resizebox{\columnwidth}{!}{%
\begin{tabular}{l|cc|cc||cc}
\hline
\multirow{2}{*}{\textbf{Dataset}} & \multicolumn{2}{c|}{\textbf{WildGS (Full)} \cite{zheng2025wildgs}} & \multicolumn{2}{c||}{\textbf{WildGS (Fast)} \cite{zheng2025wildgs}} & \multicolumn{2}{c}{\textbf{Ours}} \\ \cline{2-7} 
 & FPS $\uparrow$ & ATE $\downarrow$ & FPS $\uparrow$ & ATE $\downarrow$ & \textbf{FPS} $\uparrow$ & \textbf{ATE} $\downarrow$ \\ \hline
Wild-SLAM & 0.49 & \textbf{0.46} & 1.96 & 0.48 & \textbf{11.24} & 2.17 \\
Bonn & 0.50 & \textbf{2.31} & 2.13 & 2.47 & \textbf{10.13} & 2.46 \\ \hline
\end{tabular}%
}
\vspace{-15pt}
\end{table}

Table~\ref{tab:bonn_dynamic} presents the quantitative evaluation on the Bonn Dynamic sequences. Our method attains an Average ATE of 2.46 cm, achieving the second-best overall performance, closely approaching the SOTA WildGS-SLAM (2.31 cm) \cite{zheng2025wildgs}.

\definecolor{firstcolor}{rgb}{0.776, 0.937, 0.808}  
\definecolor{secondcolor}{rgb}{0.902, 0.961, 0.784} 
\definecolor{thirdcolor}{rgb}{1.0, 0.922, 0.612}    

\begin{table*}[t]
\centering
\caption{Tracking Performance and Ablation Study on Bonn RGB-D Dynamic Dataset (ATE RMSE [cm]). \protect\\ Best results are highlighted as \protect\colorbox{firstcolor}{\textbf{first}}, \protect\colorbox{secondcolor}{second}, and \protect\colorbox{thirdcolor}{third}. ‘F’ indicates tracking failure.}
\label{tab:bonn_dynamic}
\resizebox{\textwidth}{!}{%
\begin{tabular}{lcccccccc|c}
\hline
Method & Balloon & Balloon2 & Crowd & Crowd2 & Person & Person2 & Moving & Moving2 & Avg. \\ \hline
DSO \cite{engel2017direct} & 7.3 & 21.8 & 10.1 & 7.6 & 30.6 & 26.5 & 4.7 & 11.2 & 15.0 \\
DROID-SLAM \cite{teed2021droid} & 7.5 & 4.1 & 5.2 & 6.5 & 4.3 & 5.4 & 2.3 & 4.0 & 4.91 \\
MonoGS \cite{matsuki2024gaussian} & 15.3 & 17.3 & 11.3 & 7.3 & 26.4 & 35.2 & 22.2 & 47.2 & 22.8 \\
Splat-SLAM \cite{sandstrom2025splat} & 8.8 & 3.0 & 6.8 & F & 4.9 & 25.8 & 1.7 & 3.0 & - \\
DynaMoN (MS) \cite{schischka2024dynamon} & 6.8 & 3.8 & 6.1 & 5.6 & \cellcolor{firstcolor}\textbf{2.4} & 3.5 & 1.4 & 2.6 & 4.02 \\
DynaMoN (MS\&SS) \cite{schischka2024dynamon} & 2.8 & 2.7 & 3.5 & \cellcolor{thirdcolor}2.8 & 14.8 & \cellcolor{firstcolor}\textbf{2.2} & \cellcolor{thirdcolor}1.3 & 2.7 & 4.10 \\
MonST3R-SW \cite{zhang2024monst3r} & 5.4 & 7.2 & 5.4 & 6.9 & 11.9 & 11.1 & 3.3 & 7.4 & 7.3 \\
MegaSaM \cite{li2025megasam} & 3.7 & \cellcolor{thirdcolor}2.6 & \cellcolor{secondcolor}1.6 & 7.2 & 4.1 & 4.0 & 1.4 & 3.4 & 3.51 \\
WildGS-SLAM \cite{zheng2025wildgs} & 2.8 & \cellcolor{secondcolor}2.4 & \cellcolor{firstcolor}\textbf{1.5} & \cellcolor{secondcolor}2.3 & 3.1 & \cellcolor{secondcolor}2.7 & 1.6 & \cellcolor{thirdcolor}2.2 & \cellcolor{firstcolor}\textbf{2.31} \\ \hline
\multicolumn{10}{l}{\textit{Ablation Study}} \\ \hline
w/o Conf Reweighting & \cellcolor{thirdcolor}2.5 & \cellcolor{firstcolor}\textbf{1.7} & 11.0 & 3.3 & \cellcolor{thirdcolor}2.9 & 3.5 & \cellcolor{secondcolor}1.0 & \cellcolor{secondcolor}1.7 & \cellcolor{thirdcolor}3.45 \\
w/o All-Tracker & \cellcolor{secondcolor}2.0 & \cellcolor{firstcolor}\textbf{1.7} & 8.3 & 6.4 & 3.2 & \cellcolor{thirdcolor}3.2 & 1.6 & 2.7 & 3.64 \\ 
\textbf{All-St3R SLAM (Ours)} & \cellcolor{firstcolor}\textbf{1.5} & \cellcolor{secondcolor}2.4 & \cellcolor{thirdcolor}2.7 & \cellcolor{firstcolor}\textbf{2.1} & \cellcolor{secondcolor}2.6 & 6.3 & \cellcolor{firstcolor}\textbf{0.6} & \cellcolor{firstcolor}\textbf{1.5} & \cellcolor{secondcolor}2.46 \\ \hline
\end{tabular}%
}
\vspace{-5pt}
\end{table*}

Table~\ref{tab:tum_dynamic} presents the evaluation on the TUM RGB-D Dynamic sequences. While some recent learning-based methods such as MonoGS (26.4 cm) \cite{matsuki2024gaussian} and MonST3R-SW (20.8 cm) \cite{zhang2024monst3r} suffer from catastrophic failures or large drift in these highly dynamic scenarios, our method maintains a stable and competitive performance with an average ATE of 2.27 cm, operating at \textbf{$\sim$10.18 FPS} with a peak GPU memory of 9.74 GB.

\begin{table*}[t]
\centering
\caption{Tracking Performance and Ablation Study on TUM RGB-D Dynamic Dataset (ATE RMSE [cm]). \protect\\ Best results are highlighted as \protect\colorbox{firstcolor}{\textbf{first}}, \protect\colorbox{secondcolor}{second}, and \protect\colorbox{thirdcolor}{third}. Results are marked with '-' when the released implementation could not be successfully reproduced.}
\label{tab:tum_dynamic}
\footnotesize 
\setlength{\tabcolsep}{6pt}
\renewcommand{\arraystretch}{0.75}
\resizebox{\textwidth}{!}{%
\begin{tabular}{l|ccccccccc|c}
\hline
Method & f2/dp & f3/ss & f3/sx & f3/sr & f3/shs & f3/ws & f3/wr & f3/wx & f3/whs & Avg. \\ \hline
DSO \cite{engel2017direct} & 2.2 & 1.7 & 11.5 & 3.7 & 12.4 & 1.5 & 12.9 & 13.8 & 40.7 & 11.1 \\
DROID-SLAM \cite{teed2021droid} & \cellcolor{firstcolor}\textbf{0.6} & \cellcolor{firstcolor}\textbf{0.5} & \cellcolor{secondcolor}0.9 & \cellcolor{secondcolor}2.2 & \cellcolor{firstcolor}\textbf{1.4} & 1.2 & \cellcolor{thirdcolor}1.6 & \cellcolor{thirdcolor}4.0 & 2.2 & \cellcolor{secondcolor}1.62 \\
MonoGS \cite{matsuki2024gaussian} & 112.8 & 1.2 & 6.1 & 5.1 & 28.3 & 1.1 & 21.5 & 17.4 & 44.2 & 26.4 \\
Splat-SLAM \cite{sandstrom2025splat} & \cellcolor{secondcolor}0.7 & \cellcolor{firstcolor}\textbf{0.5} & \cellcolor{secondcolor}0.9 & \cellcolor{thirdcolor}2.3 & \cellcolor{secondcolor}1.5 & 2.3 & \cellcolor{firstcolor}\textbf{1.3} & \cellcolor{secondcolor}3.9 & 2.2 & 1.71 \\
DynaMoN (MS) \cite{schischka2024dynamon} & \cellcolor{firstcolor}\textbf{0.6} & \cellcolor{firstcolor}\textbf{0.5} & \cellcolor{secondcolor}0.9 & \cellcolor{firstcolor}\textbf{2.1} & \cellcolor{thirdcolor}1.9 & 1.4 & \cellcolor{secondcolor}1.4 & \cellcolor{secondcolor}3.9 & \cellcolor{thirdcolor}2.0 & \cellcolor{thirdcolor}1.63 \\
DynaMoN (MS\&SS) \cite{schischka2024dynamon} & \cellcolor{secondcolor}0.7 & \cellcolor{firstcolor}\textbf{0.5} & \cellcolor{secondcolor}0.9 & 2.4 & 2.3 & \cellcolor{thirdcolor}0.7 & \cellcolor{secondcolor}1.4 & \cellcolor{secondcolor}3.9 & \cellcolor{secondcolor}1.9 & \cellcolor{thirdcolor}1.63 \\
DDN-SLAM (RGB) \cite{li2025ddn} & - & - & \cellcolor{thirdcolor}1.3 & - & 3.1 & 2.5 & 2.8 & 8.9 & 4.1 & - \\
MonST3R-SW \cite{zhang2024monst3r} & 51.6 & 2.4 & 28.2 & 5.4 & 36.5 & 2.2 & 27.3 & 13.6 & 19.8 & 20.8 \\
WildGS-SLAM \cite{zheng2025wildgs} & \cellcolor{thirdcolor}1.4 & \cellcolor{firstcolor}\textbf{0.5} & \cellcolor{firstcolor}\textbf{0.8} & 2.4 & 2.0 & \cellcolor{firstcolor}\textbf{0.4} & \cellcolor{firstcolor}\textbf{1.3} & \cellcolor{firstcolor}\textbf{3.3} & \cellcolor{firstcolor}\textbf{1.6} & \cellcolor{firstcolor}\textbf{1.51} \\ \hline
\multicolumn{11}{l}{\textit{Ablation Study}} \\ \hline
w/o Conf Reweighting & 2.3 & \cellcolor{thirdcolor}0.8 & 1.5 & 3.1 & 5.4 & 1.0 & 1.9 & 5.0 & 3.4 & 2.71 \\
w/o All-Tracker & \cellcolor{secondcolor}0.7 & \cellcolor{secondcolor}0.6 & 1.5 & 2.4 & 3.2 & 1.5 & 2.4 & 6.6 & 4.2 & 2.57 \\ 
\textbf{All-St3R SLAM (Ours)} & 3.0 & 0.9 & \cellcolor{thirdcolor}1.3 & \cellcolor{firstcolor}\textbf{2.1} & 2.2 & \cellcolor{secondcolor}0.6 & 2.1 & \cellcolor{firstcolor}\textbf{3.3} & 4.9 & 2.27 \\ \hline
\end{tabular}%
}
\vspace{-10pt}
\end{table*}



\begin{figure}
\centering
\includegraphics[width=\columnwidth]{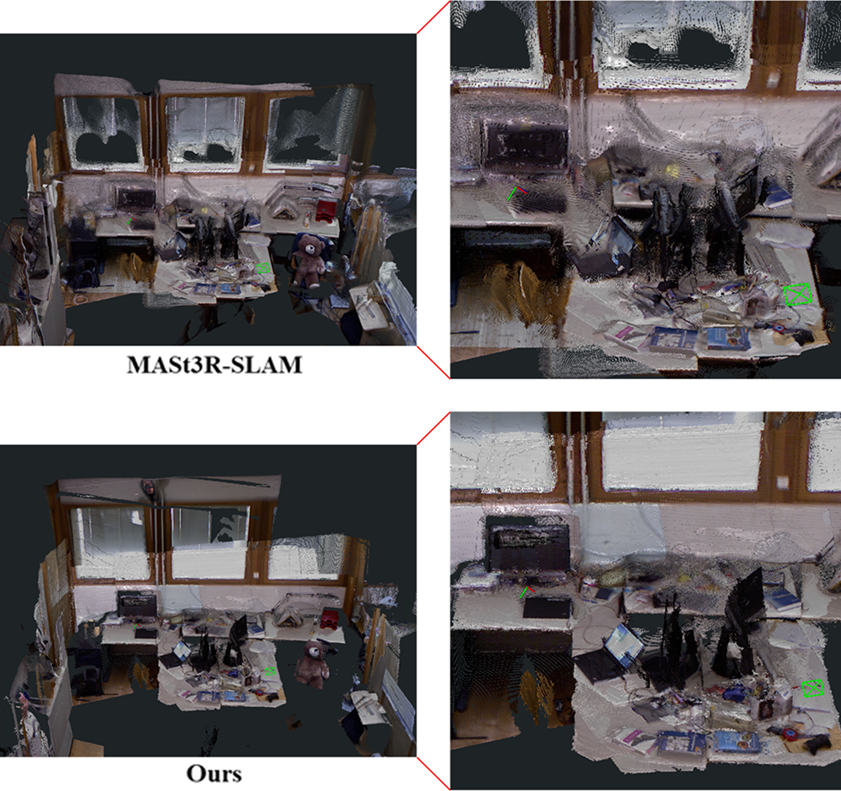}
\caption{\textbf{Qualitative comparison on a Static Scene.} Reconstruction results on \textit{freiburg1\_room}. Even without the heavy matching head of MASt3R~\cite{leroy2024grounding}, our method (combining DUSt3R\cite{wang2024dust3r} with All-Tracker\cite{harley2025alltracker}) exhibits superior geometric alignment in the reconstructed map. This confirms that our lightweight flow integration effectively compensates for the absence of learned matching features.}
\label{fig:static_comparison}
\vspace{-5pt}
\end{figure}

Our system also maintains high accuracy in static environments. Despite lacking MASt3R's learned matching head, our combination of DUSt3R and All-Tracker yields reconstruction quality that is visually comparable to MASt3R-SLAM \cite{murai2025mast3r}, with occasional improvements in several scenes (Fig.~\ref{fig:static_comparison}). Quantitatively, our average ATE on the TUM static sequences is 0.8 cm higher than MASt3R-SLAM (3.0 vs 3.8 cm), confirming that our dynamic-handling design does not significantly degrade foundational tracking stability.

\begin{figure}
\centering
\includegraphics[width=\columnwidth]{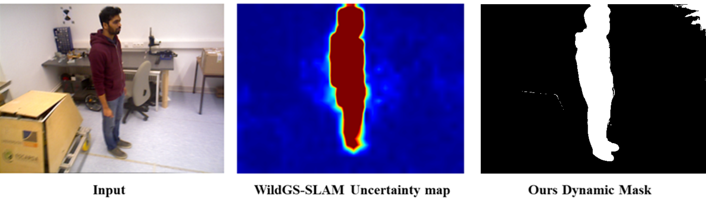}
\caption{\textbf{Mask Quality Comparison.} Middle: Uncertainty map from WildGS-SLAM \cite{zheng2025wildgs}, which is blurry due to feature downsampling. Right: Our pixel-wise dynamic mask on the Bonn dataset, showing high-resolution sharpness and better boundary adherence.}
\label{fig:wildgs_comparison}
\vspace{-15pt}
\end{figure}

\subsection{Dynamic Masking Analysis}
\label{sec:exp_masking}

\begin{figure}
\vspace{-10pt}
\centering
\includegraphics[width=\columnwidth]{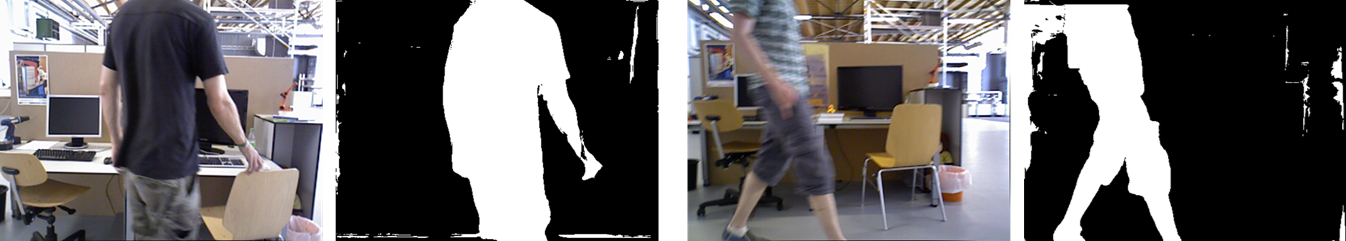}
\caption{\textbf{KeyFrame Dynamic Masks.} Visualization of dynamic masks on keyframes from \textit{freiburg3\_walking\_xyz}. The system effectively identifies the moving person.}
\label{fig:dynamic_mask_tum}
\vspace{-10pt}
\end{figure}

\begin{figure}[h]
\centering
\includegraphics[width=\columnwidth]{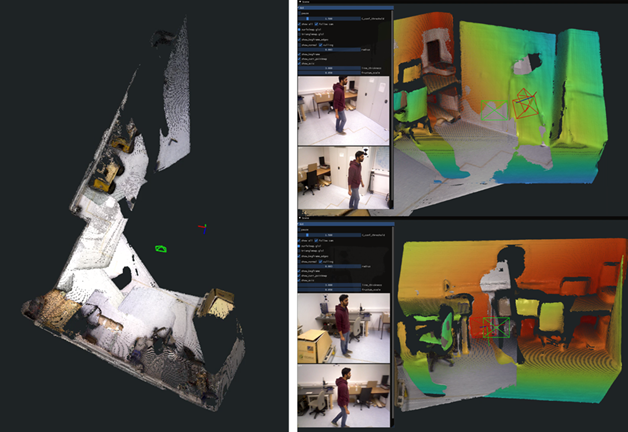}
\caption{\textbf{Reconstruction on Bonn Dataset.} Left: Final reconstruction of \textit{person\_tracking}. The moving person is completely removed, resulting in clean geometry. Right: Live tracking visualization showing the current frame (rainbow) aligned to the canonical map, with the dynamic person effectively masked out.}
\label{fig:bonn_visualize}
\vspace{-5pt}
\end{figure}

Existing methods like WildGS-SLAM\cite{zheng2025wildgs} utilize uncertainty maps derived from finetuned DINOv2 features. As noted in their work \cite{zheng2025wildgs}, these maps often appear blurry because features are extracted at low resolution (1/14 of original).
Fig.~\ref{fig:wildgs_comparison} contrasts this with our approach. While the uncertainty map (Left) is coarse, our dynamic mask (Right) is sharp at full resolution.
Fig.~\ref{fig:dynamic_mask_tum} visualizes the dynamic masks on keyframes from the \textit{freiburg3\_walking\_xyz} sequence. Our system successfully isolates the moving person.
The effectiveness of our dynamic masking is further reflected in the resulting scene reconstruction. Fig.~\ref{fig:bonn_visualize} presents qualitative results on the Bonn dataset. 
As shown in the figure, 
the moving person is successfully removed from the reconstructed geometry, producing a clean and stable map. During live tracking, the current frame 
remains well aligned with the canonical map while the dynamic person is consistently masked out, demonstrating that the proposed pixel-wise masking effectively prevents dynamic objects from corrupting the map.

\begin{figure}[h]
\centering
\includegraphics[width=\columnwidth]{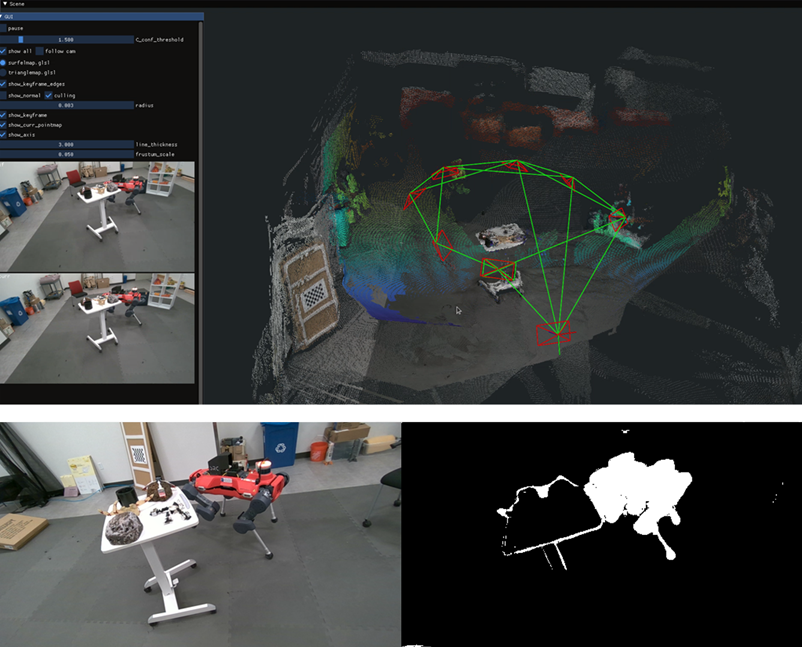}
\caption{\textbf{Generalization: Quadruped Robot.} Top: Clean reconstruction. Bottom: Dynamic mask. Our method successfully detects and removes the ANYmal robot without prior knowledge of the object class.}
\label{fig:generalization_robot}
\vspace{-20pt}
\end{figure}



\subsection{Generalizable Dynamic Mask}
\label{sec:exp_generalization}

A significant advantage of our approach is its generalization to unseen objects that pretrained networks may miss.
Fig.~\ref{fig:generalization_robot} demonstrates this on the \textit{Wild-SLAM MoCap Dataset(scene2/ANYmal2)} dataset. Our system successfully masks the quadruped robot—a non-standard class for many object detectors—solely based on motion cues. This result confirms that our method generalizes well to arbitrary dynamic environments.


\subsection{Ablation Study}


To evaluate the individual contributions of each core component in our framework, we conduct an ablation study on the Bonn and TUM RGB-D datasets. We compare our full pipeline (\textbf{All-St3R SLAM}) against two configurations: \textit{w/o Conf Reweighting}, and \textit{w/o All-Tracker}. As shown in Table~\ref{tab:bonn_dynamic} and Table~\ref{tab:tum_dynamic}, our full method provides more stable tracking in highly dynamic scenarios. Particularly in complex environments containing multiple dynamic objects—such as the \textit{Crowd} and \textit{Crowd2} sequences of the Bonn dataset—the other two configurations experience a notable increase in trajectory error, whereas our method remains robust. 

This performance gap, especially in the \textit{w/o All-Tracker} variant, highlights the inherent limitation of relying solely on DUSt3R's geometric matching ($\mathcal{M}_{geo}$). Because $\mathcal{M}_{geo}$ computes correspondences strictly based on the spatial proximity of 3D points without appearance cues, it is highly vulnerable in crowded scenes with co-occurring moving entities or dynamic occlusions. In such scenarios, the generated point clouds become structurally intertwined, causing $\mathcal{M}_{geo}$ to establish false matches between distinct dynamic entities (e.g., different moving persons) or between dynamic points and the static background simply due to their spatial closeness. In contrast, All-Tracker tracks dense correspondences purely via image-level visual appearance, bypassing these geometric proximity traps to ensure stable tracking. 

Conversely, in sequences with extensive static regions, our full pipeline shows comparable or slightly lower performance. 
This minor degradation occurs because the confidence reweighting strategy unnecessarily adjusts confidence values, concentrating camera pose alignment on specific regions even in the absence of dynamic motion. Moreover, the dynamic masking module removes useful static points that are otherwise preserved in the \textit{w/o All-Tracker} variant.



\section{DISCUSSION \& LIMITATIONS}

Although our framework demonstrates robust tracking in diverse dynamic environments, certain limitations remain. 

First, since DUSt3R does not provide absolute metric scale, if a real-world metric scale is required, it can be achieved by introducing a scale alignment module or integrating metric priors from an IMU to anchor the initial scale.

Second, while our motion discrepancy check filters out moving objects regardless of their speed, a vulnerability can occur in slow-moving scenarios when a new keyframe is created. If an object moves slowly and a sudden change in camera motion triggers a new keyframe before the object is completely masked, unmasked residuals will remain in the map. These residuals will not be removed and will cause map corruption unless a loop closure occurs near that viewpoint. 
Furthermore, increasing the subsampling factor is not always beneficial because it reduces the overlapping static background region due to large baseline changes, ultimately leading to camera tracking failure.

Lastly, DUSt3R occasionally reconstructs a dynamic object as a flat texture on the static background (e.g., as if printed on a wall). This lack of geometric separation prevents the masking module from detecting sufficient motion discrepancy, failing to filter out the dynamic object.

\section{CONCLUSIONS}
In this work, we presented a real-time monocular dense SLAM framework capable of robust operation in dynamic environments without relying on computationally expensive priors or depth sensor inputs. 
By synthesizing the dense geometric capabilities of DUSt3R~\cite{wang2024dust3r} with our lightweight version of All-Tracker~\cite{harley2025alltracker}, we introduced a \textit{dense, geometry-driven} approach to dynamic masking.

Our core contribution lies in the pixel-wise motion discrepancy check, which identifies independent object motion by comparing the pixel motion predicted by ego-motion with the observed point tracking motion. This strategy addresses the limitations of existing object-level masking and hybrid methods: it minimizes over-removal by preserving static regions within moving objects and generalizes effortlessly to unseen object categories.

Ultimately,
our system significantly outperforms existing dense dynamic SLAM systems in speed while maintaining competitive trajectory accuracy.



\section*{ACKNOWLEDGMENT}
This work was supported by IITP grant funded by the Korea government(MSIT)(No.RS-2026-25516746, Development of On-Device VLA-Based Real-Time Action Intelligence Technology for Task Execution of Autonomous Agents) and the Korea Institute of Science and Technology (KIST) Institutional Program (26E0061).

\balance




\bibliographystyle{IEEEtran}
\bibliography{refs}

@article{cadena2017past,
  title={Past, present, and future of simultaneous localization and mapping: Toward the robust-perception age},
  author={Cadena, Cesar and Carlone, Luca and Carrillo, Henry and Latif, Yasir and Scaramuzza, Davide and Neira, Jos{\'e} and Reid, Ian and Leonard, John J},
  journal={IEEE Transactions on robotics},
  volume={32},
  number={6},
  pages={1309--1332},
  year={2017},
  publisher={IEEE}
}

@article{duan2022survey,
  title={A survey of embodied ai: From simulators to research tasks},
  author={Duan, Jiafei and Yu, Samson and Tan, Hui Li and Zhu, Hongyuan and Tan, Cheston},
  journal={IEEE Transactions on Emerging Topics in Computational Intelligence},
  volume={6},
  number={2},
  pages={230--244},
  year={2022},
  publisher={IEEE}
}

@inproceedings{murai2025mast3r,
  title={MASt3R-SLAM: Real-time dense SLAM with 3D reconstruction priors},
  author={Murai, Riku and Dexheimer, Eric and Davison, Andrew J},
  booktitle={Proceedings of the Computer Vision and Pattern Recognition Conference},
  pages={16695--16705},
  year={2025}
}

@article{zhang2020vdo,
  title={VDO-SLAM: A visual dynamic object-aware SLAM system},
  author={Zhang, Jun and Henein, Mina and Mahony, Robert and Ila, Viorela},
  journal={arXiv preprint arXiv:2005.11052},
  year={2020}
}

@article{yang2019cubeslam,
  title={Cubeslam: Monocular 3-d object slam},
  author={Yang, Shichao and Scherer, Sebastian},
  journal={IEEE Transactions on Robotics},
  volume={35},
  number={4},
  pages={925--938},
  year={2019},
  publisher={IEEE}
}

@inproceedings{tan2013robust,
  title={Robust monocular SLAM in dynamic environments},
  author={Tan, Wei and Liu, Haomin and Dong, Zilong and Zhang, Guofeng and Bao, Hujun},
  booktitle={2013 IEEE International Symposium on Mixed and Augmented Reality (ISMAR)},
  pages={209--218},
  year={2013},
  organization={IEEE}
}

@article{bescos2018dynaslam,
  title={DynaSLAM: Tracking, mapping, and inpainting in dynamic scenes},
  author={Bescos, Berta and F{\'a}cil, Jos{\'e} M and Civera, Javier and Neira, Jos{\'e}},
  journal={IEEE robotics and automation letters},
  volume={3},
  number={4},
  pages={4076--4083},
  year={2018},
  publisher={IEEE}
}

@inproceedings{lin2014microsoft,
  title={Microsoft coco: Common objects in context},
  author={Lin, Tsung-Yi and Maire, Michael and Belongie, Serge and Hays, James and Perona, Pietro and Ramanan, Deva and Doll{\'a}r, Piotr and Zitnick, C Lawrence},
  booktitle={European conference on computer vision},
  pages={740--755},
  year={2014},
  organization={Springer}
}

@article{long2025dog,
  title={DOG-SLAM: Enhancing dynamic visual SLAM precision through GMM-based dynamic object removal and ORB-boost},
  author={Long, Jiming and Wang, Fei and Liu, Mingyang and Wang, Yu and Zou, Qiang},
  journal={IEEE Transactions on Instrumentation and Measurement},
  year={2025},
  publisher={IEEE}
}

@inproceedings{zheng2025wildgs,
  title={Wildgs-slam: Monocular gaussian splatting slam in dynamic environments},
  author={Zheng, Jianhao and Zhu, Zihan and Bieri, Valentin and Pollefeys, Marc and Peng, Songyou and Armeni, Iro},
  booktitle={Proceedings of the Computer Vision and Pattern Recognition Conference},
  pages={11461--11471},
  year={2025}
}

@article{oquab2023dinov2,
  title={Dinov2: Learning robust visual features without supervision},
  author={Oquab, Maxime and Darcet, Timoth{\'e}e and Moutakanni, Th{\'e}o and Vo, Huy and Szafraniec, Marc and Khalidov, Vasil and Fernandez, Pierre and Haziza, Daniel and Massa, Francisco and El-Nouby, Alaaeldin and others},
  journal={arXiv preprint arXiv:2304.07193},
  year={2023}
}

@article{kerbl20233d,
  title={3D Gaussian splatting for real-time radiance field rendering.},
  author={Kerbl, Bernhard and Kopanas, Georgios and Leimk{\"u}hler, Thomas and Drettakis, George},
  journal={ACM Trans. Graph.},
  volume={42},
  number={4},
  pages={139--1},
  year={2023}
}

@inproceedings{wang2024dust3r,
  title={Dust3r: Geometric 3d vision made easy},
  author={Wang, Shuzhe and Leroy, Vincent and Cabon, Yohann and Chidlovskii, Boris and Revaud, Jerome},
  booktitle={Proceedings of the IEEE/CVF Conference on Computer Vision and Pattern Recognition},
  pages={20697--20709},
  year={2024}
}

@inproceedings{harley2025alltracker,
  title={Alltracker: Efficient dense point tracking at high resolution},
  author={Harley, Adam W and You, Yang and Sun, Xinglong and Zheng, Yang and Raghuraman, Nikhil and Gu, Yunqi and Liang, Sheldon and Chu, Wen-Hsuan and Dave, Achal and You, Suya and others},
  booktitle={Proceedings of the IEEE/CVF International Conference on Computer Vision},
  pages={5253--5262},
  year={2025}
}

@inproceedings{leroy2024grounding,
  title={Grounding image matching in 3d with mast3r},
  author={Leroy, Vincent and Cabon, Yohann and Revaud, J{\'e}r{\^o}me},
  booktitle={European Conference on Computer Vision},
  pages={71--91},
  year={2024},
  organization={Springer}
}

@inproceedings{matsuki2024gaussian,
  title={Gaussian splatting slam},
  author={Matsuki, Hidenobu and Murai, Riku and Kelly, Paul HJ and Davison, Andrew J},
  booktitle={Proceedings of the IEEE/CVF Conference on Computer Vision and Pattern Recognition},
  pages={18039--18048},
  year={2024}
}

@inproceedings{sandstrom2025splat,
  title={Splat-slam: Globally optimized rgb-only slam with 3d gaussians},
  author={Sandstr{\"o}m, Erik and Zhang, Ganlin and Tateno, Keisuke and Oechsle, Michael and Niemeyer, Michael and Zhang, Youmin and Patel, Manthan and Van Gool, Luc and Oswald, Martin and Tombari, Federico},
  booktitle={Proceedings of the Computer Vision and Pattern Recognition Conference},
  pages={1680--1691},
  year={2025}
}

@article{li2025ddn,
  title={Ddn-slam: Real time dense dynamic neural implicit slam},
  author={Li, Mingrui and Guo, Zhetao and Deng, Tianchen and Zhou, Yiming and Ren, Yuxiang and Wang, Hongyu},
  journal={IEEE Robotics and Automation Letters},
  year={2025},
  publisher={IEEE}
}

@article{engel2017direct,
  title={Direct sparse odometry},
  author={Engel, Jakob and Koltun, Vladlen and Cremers, Daniel},
  journal={IEEE transactions on pattern analysis and machine intelligence},
  volume={40},
  number={3},
  pages={611--625},
  year={2017},
  publisher={IEEE}
}

@article{teed2021droid,
  title={Droid-slam: Deep visual slam for monocular, stereo, and rgb-d cameras},
  author={Teed, Zachary and Deng, Jia},
  journal={Advances in neural information processing systems},
  volume={34},
  pages={16558--16569},
  year={2021}
}

@article{schischka2024dynamon,
  title={Dynamon: Motion-aware fast and robust camera localization for dynamic neural radiance fields},
  author={Schischka, Nicolas and Schieber, Hannah and Karaoglu, Mert Asim and Gorgulu, Melih and Gr{\"o}tzner, Florian and Ladikos, Alexander and Navab, Nassir and Roth, Daniel and Busam, Benjamin},
  journal={IEEE Robotics and Automation Letters},
  year={2024},
  publisher={IEEE}
}

@article{zhang2024monst3r,
  title={Monst3r: A simple approach for estimating geometry in the presence of motion},
  author={Zhang, Junyi and Herrmann, Charles and Hur, Junhwa and Jampani, Varun and Darrell, Trevor and Cole, Forrester and Sun, Deqing and Yang, Ming-Hsuan},
  journal={arXiv preprint arXiv:2410.03825},
  year={2024}
}

@inproceedings{li2025megasam,
  title={MegaSaM: Accurate, fast and robust structure and motion from casual dynamic videos},
  author={Li, Zhengqi and Tucker, Richard and Cole, Forrester and Wang, Qianqian and Jin, Linyi and Ye, Vickie and Kanazawa, Angjoo and Holynski, Aleksander and Snavely, Noah},
  booktitle={Proceedings of the Computer Vision and Pattern Recognition Conference},
  pages={10486--10496},
  year={2025}
}

@article{ravi2024sam2,
  title={SAM 2: Segment Anything in Images and Videos},
  author={Ravi, Nikhila and Gabeur, Valentin and Hu, Yuan-Ting and Hu, Ronghang and Ryali, Chaitanya and Ma, Tengyu and Khedr, Haitham and R{\"a}dle, Roman and Rolland, Chloe and Gustafson, Laura and Mintun, Eric and Pan, Junting and Alwala, Kalyan Vasudev and Carion, Nicolas and Wu, Chao-Yuan and Girshick, Ross and Doll{\'a}r, Piotr and Feichtenhofer, Christoph},
  journal={arXiv preprint arXiv:2408.00714},
  url={https://arxiv.org/abs/2408.00714},
  year={2024}
}

@inproceedings{chen2025easi3r,
  title={Easi3r: Estimating disentangled motion from dust3r without training},
  author={Chen, Xingyu and Chen, Yue and Xiu, Yuliang and Geiger, Andreas and Chen, Anpei},
  booktitle={Proceedings of the IEEE/CVF International Conference on Computer Vision},
  pages={9158--9168},
  year={2025}
}

\end{document}